\pdfoutput=1

\documentclass[11pt]{article}

\usepackage{EMNLP2022}

\usepackage{times}
\usepackage{latexsym}
\usepackage{enumitem}
\usepackage{enumerate}
\usepackage[T1]{fontenc}

\usepackage[utf8]{inputenc}

\usepackage{microtype}

\usepackage{inconsolata}

\usepackage{multirow}
\usepackage{graphics}
\usepackage{array,graphicx}
\usepackage{float}
\usepackage{arydshln}
\usepackage{xcolor}
\usepackage{soul}
\usepackage{url}
\usepackage{hyperref}
\hypersetup{colorlinks=true}

\title{Making Science Simple: Corpora for the Lay Summarisation of Scientific Literature}

\author{Tomas Goldsack$^{1}$, Zhihao Zhang$^{2}$, Chenghua Lin$^{1}$\footnotemark[1]~, Carolina Scarton$^{1}$ \\
        $^{1}$Department of Computer Science, University of Sheffield, UK \\ 
        $^{2}$School of Economics and Management, Beihang University, China\\
        \texttt{\{tgoldsack1, c.lin, c.scarton\}@sheffield.ac.uk}\\
        \texttt{zhhzhang@buaa.edu.cn}}

\begin{document}
\maketitle

\footnotetext[1]{$^{*}$Corresponding author.}

\begin{abstract}
Lay summarisation aims to jointly summarise and simplify a given text, thus making its content more comprehensible to non-experts.
Automatic approaches for lay summarisation can provide significant value in broadening access to scientific literature, enabling a greater degree of both interdisciplinary knowledge sharing and public understanding when it comes to research findings. However, current corpora for this task are limited in their size and scope, hindering the development of broadly applicable data-driven approaches. 
Aiming to rectify these issues, we present two novel lay summarisation datasets, PLOS (large-scale) and eLife (medium-scale), each of which contains biomedical journal articles alongside expert-written lay summaries.
We provide a thorough characterisation of our lay summaries, 
highlighting differing levels of readability and abstractiveness
between datasets that can be leveraged to support the needs of different applications.
Finally, we benchmark our datasets using mainstream summarisation approaches and perform a manual evaluation with domain experts, demonstrating their utility and casting light on the key challenges of this task. 
Our code and datasets are available at \url{https://github.com/TGoldsack1/Corpora_for_Lay_Summarisation}.
\end{abstract}

\section{Introduction}

Scientific publications contain information that is essential for the preservation and progression of our understanding across all scientific disciplines. 
Typically being highly technical in nature, such articles tend to assume a degree of background knowledge and make use of domain-specific language, \textcolor{black}{making them difficult to comprehend for one lacking the required expertise (i.e., a \textit{lay person}). These factors often limit the impact of research to only its direct community \citep{albert2015, albert2021} and, more dangerously, can cause readers (members of the public, journalists, etc.) to misinterpret research findings \citep{kuehne2015}.}
\begin{figure}[t]
    \noindent\fbox{%
        \parbox{\columnwidth}{%
            \textbf{Technical Abstract} \\
            \begingroup
                \fontsize{10pt}{12pt}\selectfont
                The virus SARS-CoV-2 can exploit biological vulnerabilities (e.g. host proteins) in susceptible hosts that predispose to the development of severe COVID-19. To identify host proteins that may contribute to the risk of severe COVID-19, we undertook proteome-wide genetic colocalisation tests, and polygenic (pan) and cis-Mendelian randomisation analyses leveraging publicly available protein and COVID-19 datasets...
            \endgroup \vspace{4pt}
            
            \textbf{Lay Summary} \\
            \begingroup
                \fontsize{10pt}{12pt}\selectfont
                Individuals who become infected with the virus that causes COVID-19 can experience a wide variety of symptoms. These can range from no symptoms or minor symptoms to severe illness and death. Key demographic factors, such as age, gender and race, are known to affect how susceptible an individual is to infection. However, molecular factors, such as unique gene mutations and gene expression levels can also have a major impact on patient responses by affecting the levels of proteins in the body...
            \endgroup
        }
    } 
    \caption{The first few sentences of the abstract and lay summary of an eLife article, illustrating differences in the language and focus on background information.}
    \label{fig:PLOS_example}
\end{figure}

This latter point is especially important for biomedical research which, \textcolor{black}{in addition to having particularly dynamic and confusing terminology \citep{SMITH2006288,peng2021named}}, has the potential to directly impact people's decision-making regarding health-related issues, with a pertinent example of this being the widespread misinformation seen during the COVID-19 pandemic \citep{Islam2020}.
Aiming to address these challenges, some academic journals 
\textcolor{black}{choose to publish \textit{lay summaries} that
clearly and concisely explain the context and significance of an article using non-specialist language. Figure \ref{fig:PLOS_example} illustrates how simplifying jargon (e.g., ``SARS-CoV-2" $\rightarrow$ ``the virus that causes COVID-19") and focusing on background information allows a reader to better understand a complex scientific topic.}
However, in addition to placing an extra burden on authors, lay summaries are not yet ubiquitous and focus only on newly published articles.

Automatic text summarisation can provide significant value in the generation of scientific lay summaries. 
Although previous use of summarisation techniques for scientific articles has largely focused on generating a technical summary (e.g., the abstract), 
\textcolor{black}{only a few} have addressed the task of lay summarisation and introduced datasets to facilitate its study \citep{Chandrasekaran2020-df, Guo2020-ba, Zaman2020-kx}. However, compared to datasets ordinarily used for training supervised summarisation models, these resources are relatively small (ranging from 572 to 6,695 articles), presenting a significant barrier to the deployment of large data-driven approaches that require training on large amounts of parallel data. 
Furthermore, these resources are somewhat fragmented in terms of their framing of the task, 
\textcolor{black}{making use of article and summary formats that limit their applicability to broader biomedical literature.} 
These factors hinder the progression of the field and the development of usable models that can be used to make scientific content accessible to a wider audience.

To help alleviate these issues, we introduce \textbf{two new datasets} derived from different academic journals \textbf{within the biomedical domain} - PLOS and eLife \textcolor{black}{(\S \ref{sec:our-datasets}}).
Both datasets use the full journal article as the source, enabling the training of models which can be broadly applied to wider literature.  
PLOS is significantly larger than currently available datasets and makes use of short author-written lay summaries (150-200 words), whereas eLife's summaries are approximately twice as long and written by expert editors who are well-practiced in the simplification of scientific content. \textcolor{black}{Given these differences in authorship and length, we expect the lay summaries of eLife to simplify content to a greater extent, meaning our datasets are able to cater to different audiences and applications \textcolor{black}{(e.g., personalised lay summarisation)}. We confirm this via an} in-depth characterisation of the lay summaries within each dataset, quantifying ways in which they differ from the technical abstract and from each other \textcolor{black}{(\S \ref{sec:analysis})}.
\textcolor{black}{Finally, we benchmark our datasets with popular summarisation approaches using automatic metrics and conduct an expert-based manual evaluation, highlighting the utility of our datasets and key challenges for the task of lay summarisation} \textcolor{black}{(\S \ref{sec:exp}). This paper also presents a literature review (\S \ref{sec:related}), conclusions (\S \ref{sec:conclusion}), and a discussion on its limitations (\S \ref{sec:limitations}).}

\section{Related Work}
\label{sec:related}

Past attempts to automatically summarise scientific content in layman's terms have been scarce, with the most prominent example being the LaySumm subtask of 
the CL-SciSumm 2020 shared task series \citep{Chandrasekaran2020-df} which attracted a total of 8 submissions. Alongside the task, a training corpus of 572 articles and author-generated lay summaries from a multi-disciplinary collection of Elsevier-published scientific journals was provided, with submissions being evaluated on a blind test set of 37 articles. It was noted by the task organisers that the data provided was insufficient for training a model to produce a realistic lay summary.


\citet{Guo2020-ba} also make use of a single publication source to retrieve lay summaries: The Cochrane Database of Systematic Reviews (\textsc{CDSR}). Their dataset contains the abstracts of 6,695 systematic reviews paired with their respective plain-language summaries, covering various healthcare domains.
Although larger than other available datasets for lay summarisation, \textsc{CDSR} is constrained in that it only uses the abstracts of systematic reviews as source documents, and thus models trained using \textsc{CDSR} will be unlikely to generalise well to inputs that are longer than an abstract or the abstracts of other types of publication.


Alternatively, \citet{Zaman2020-kx} introduce a dataset derived from the `Eureka-Alert' science news website for the combined tasks of simplification and summarisation. Summaries consist of news articles (average length > 600 words) that aim to describe the content of a scientific publication to the non-expert. 
\textcolor{black}{However, the extensive size of reference summaries is likely to present additional challenges in model training and their news-based format limits their applicability (e.g., in automating lay summarisation for journals).}

Compared to previous resources, our datasets contain articles and lay summaries of a format that we consider to be more broadly applicable to wider literature. Additionally, PLOS is significantly larger than those currently available \textcolor{black}{(over 4$\times$ larger than CDSR)} and eLife contains summaries written by expert editors. 
\textcolor{black}{Furthermore, our work is the first to provide two datasets with different levels of readability, thus supporting the needs of different audiences and applications.}
Through each of these factors, we hope to enable the creation of more usable lay summarisation models.

\section{Our Datasets} \label{sec:our-datasets}

We introduce two datasets from different biomedical journals \textcolor{black}{(PLOS and eLife)}, each containing full scientific articles paired with manually-created lay summaries. For each data source, articles were retrieved in XML format and parsed using Python to retrieve the lay summary, abstract, and article text.\footnote{For each article, we also retrieve a number of \textit{keywords} from the meta-data, providing an indication of the high-level topics covered within the article.}  
In line with previous datasets for scientific summarisation \citep{Cohan2018-nq}, the article text is separated into sections, and the heading of each section is also retrieved. Sentences are segmented using the PySBD rule-based parser \citep{sadvilkar-neumann-2020-pysbd}, which we empirically found to outperform neural alternatives. 
We separate our datasets into training, validation, and testing splits at a ratio of 90\%/5\%/5\%. Statistics describing the contents of our datasets and that of past lay summarisation datasets are given in Table \ref{tab:datasets}.

\begin{table}[t]
    \centering
    \resizebox{\columnwidth}{!}{
        \begin{tabular}{lcccc} \hline
            \multirow{2}{*}{\textbf{Dataset}} & \multirow{2}{*}{\textbf{\# Docs}} & \multicolumn{1}{c}{\textbf{Doc}} & \multicolumn{2}{c}{\textbf{Summary}} \\   \cline{3-5}
            && \# words   & \# words  & \# sents  \\ 
                    \hline
                    LaySumm         & 572     & 4,426.1 & 82.15 & 3.8  \\
                    Eureka-Alert    & 5,204   & 5,027.0 & 635.6 & 24.3 \\
                    CDSR            & 6,695   & 576.0   & 338.2 & 16.1 \\
                    \cdashline{1-5}[1.5pt/2pt]
                    \textbf{PLOS}   & 27,525  & 5,366.7 & 175.6 & 7.8  \\
                    \textbf{eLife}  & 4,828   & 7,806.1 & 347.6 & 15.7 \\
                    \hline
        \end{tabular} 
    }
    \caption{Statistics of lay summarisation datasets, with ours given in \textbf{bold}. Words and sentences (sents) are average values.} 
    \label{tab:datasets}
\end{table}


\paragraph{PLOS} The Public Library of Science (PLOS) is an open-access publisher 
\textcolor{black}{that hosts} 
influential peer-reviewed journals across all areas of science and medicine.
Several of these journals require authors to submit an \textit{author summary} alongside their work, defined as a 150-200 word non-technical summary aimed at making the findings of a paper accessible to a wider audience, including non-scientists.\footnote{Source of PLOS author summary definition: \url{https://journals.plos.org/plosgenetics/s/submission-guidelines}}
The journals in question focus specifically on Biology, Computational Biology, Genetics, Pathogens, and Neglected Tropical Diseases.


\paragraph{eLife} eLife is an open-access peer-reviewed journal with a specific focus on biomedical and life sciences. Of the articles published in eLife, some are selected to be the subject of a  \textit{digest}, a simplified summary of the work written by expert editors based on both the article itself and questions answered by its author. Similarly to PLOS, these digests aim to explain the background and significance of a scientific article in language that is accessible to non-experts \citep{elifeDigest}.

\begin{table}[t]
    \centering
    \resizebox{\columnwidth}{!}{
        \begin{tabular}{lccccc} \hline
            \multirow{2}{*}{\textbf{Metric}} & \multicolumn{2}{c}{\textbf{Abstract}} && \multicolumn{2}{c}{\textbf{Lay Summary}} \\  \cline{2-3} \cline{5-6}
            & \textbf{PLOS} & \textbf{eLife}  && \textbf{PLOS} & \textbf{eLife} \\ 
                    \hline
                    FKGL$\downarrow$     & 15.04 & 15.57 && 14.76 & 10.92 \\
                    CLI$\downarrow$      & 16.39 & 17.68 && 15.90 & 12.51 \\
                    DCRS$\downarrow$     & 11.06 & 11.78 && 10.91 & 8.83  \\
                    WordRank$\downarrow$ & 9.08  & 9.21  && 8.98  & 8.68  \\
                    \hline
        \end{tabular} 
    }
    \caption{Mean readability scores for abstracts and lay summaries from our datasets. For all metrics, a lower score indicates greater readability.}
    \label{tab:simplicity_scores}
\end{table}

\section{Dataset Analysis} \label{sec:analysis}
We carry out several analyses comparing the lay summaries of our datasets to the respective technical abstracts. Through these analyses, we seek to highlight and quantify the key differences between these two different types of summary, as well as those present between the lay summaries of our two datasets. Specifically, we focus on readability (\S\ref{subsec:analysis-readability}), rhetorical structure (\S\ref{subsec:analysis-rhetoric}), vocabulary sharing (\S\ref{subsec:analysis-content}), and abstractiveness (\S\ref{subsec:analysis-abstractiveness}).

\begin{figure*}[t]
    \centering
    \resizebox{0.85\textwidth}{!}{%
        \includegraphics{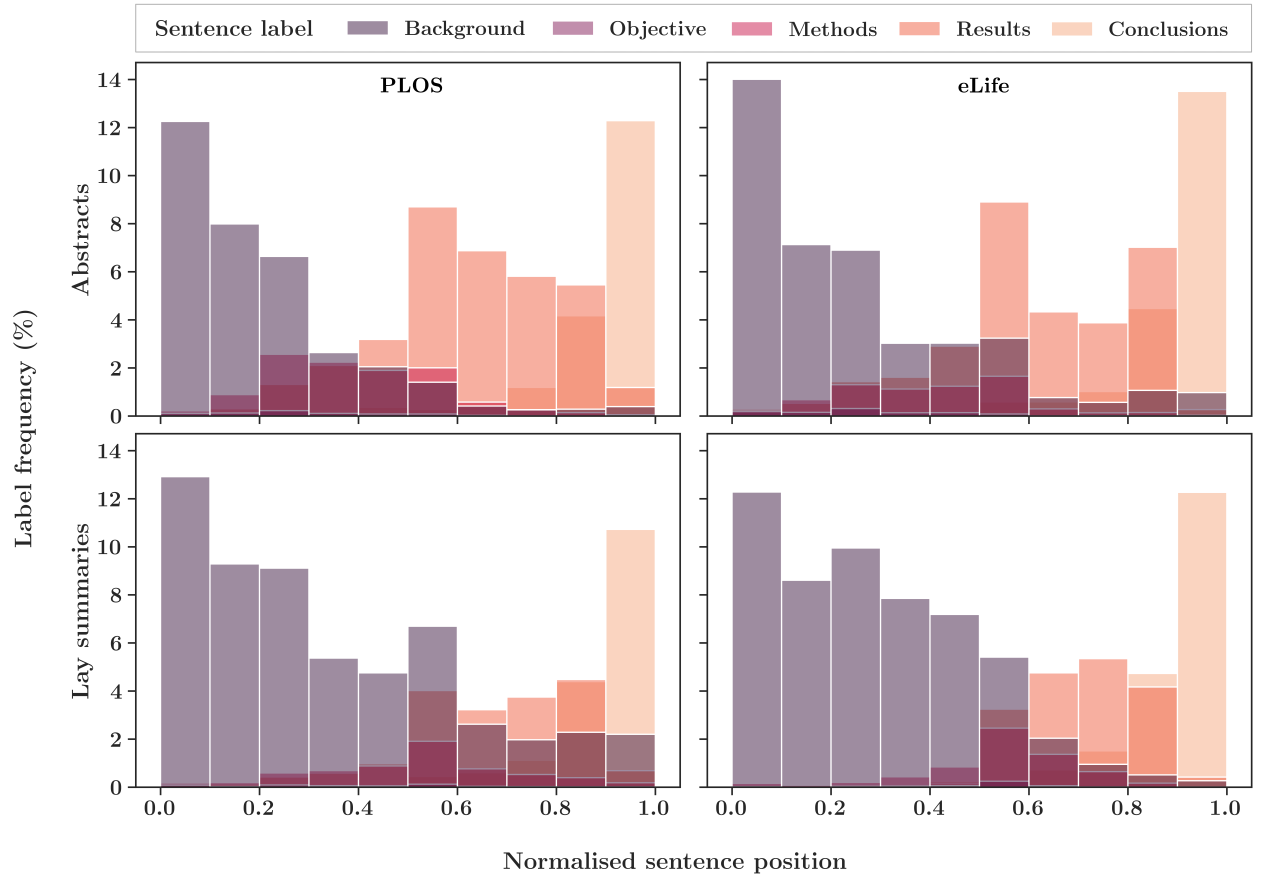}
    }
    \caption{Barplot visualising the rhetorical class distributions in our abstracts and lay summaries.}
    \label{fig:discourse_comp}
\end{figure*}

\subsection{Readability} \label{subsec:analysis-readability}
We assess the readability of our lay summaries and abstracts using several established metrics. Specifically, we employ Flesch-Kincaid Grade Level (FKGL), Coleman-Liau Index (CLI), Dale-Chall Readability Score (DCRS), and WordRank score.\footnote{Computed using the \href{https://github.com/shivam5992/textstat}{\texttt{textstat}} and \texttt{EASSE} \citep{alva-manchego-etal-2019-easse} packages.} FKGL, CLI, and DCRS provide an approximation of the (US) grade level of education required to read a given text. The formula for FKGL surrounds the total number of sentences, words, and syllables present within the text, whereas CLI is based on the number of sentences, words, and characters.  Alternatively, DCRS measures readability using the average sentence length and the number of \textit{familiar} words present, using a lookup table of the 3,000 most commonly used English words. Similarly, WordRank estimates the lexical complexity of a text based on how common the language is, using a frequency table derived from English Wikipedia.


The scores given in Table \ref{tab:simplicity_scores} show that the lay summaries of both datasets are consistently more readable than their respective abstracts across all metrics. Although these differences are small in some cases, in line with the findings of previous works \citep{Devaraj2021-ha}, we find them all to be \textbf{statistically significant} by way of Mann–Whitney U tests ($p<0.05$). 
These results indicate that lay summaries are 
\textcolor{black}{more readable} than technical abstracts in terms of both syntactic structure and lexical intelligibility.
Additionally, the lay summaries from eLife obtain lower readability scores than those of PLOS across all metrics, \textcolor{black}{confirming our expectation that they are suitable for less technical audiences.}\footnote{Manual inspection of the summaries from each dataset also support this.} 

\begin{table}[t]
    \centering
    \resizebox{\columnwidth}{!}{
    \begin{tabular}{lccccc} \hline
        \multirow{2}{*}{\textbf{Label}} & \multicolumn{2}{c}{\textbf{Abstract}} && \multicolumn{2}{c}{\textbf{Lay Summary}} \\  \cline{2-3} \cline{5-6}
        & \textbf{PLOS} & \textbf{eLife}  && \textbf{PLOS} & \textbf{eLife} \\ 
                \hline
                Background  & $35.40$ & $41.05$ && $58.11$ & $55.03$ \\
                Objective   & $0.76$  & $1.06$  && $0.54$  & $0.47$ \\
                Methods     & $10.26$ & $6.73$  && $6.24$  & $6.23$ \\
                Results     & $34.75$ & $30.60$ && $17.86$ & $18.23$ \\ 
                Conclusions & $18.83$ & $20.55$ && $17.26$ & $18.83$
                \\
                \hline
        \end{tabular}
        }
    \caption{Mean percentage of each rhetorical label within our abstracts and lay summaries.}
    \label{tab:discourse_percs}
\end{table}


\begin{figure*}[ht]
    \centering
    \resizebox{0.85\textwidth}{!}{%
        \includegraphics{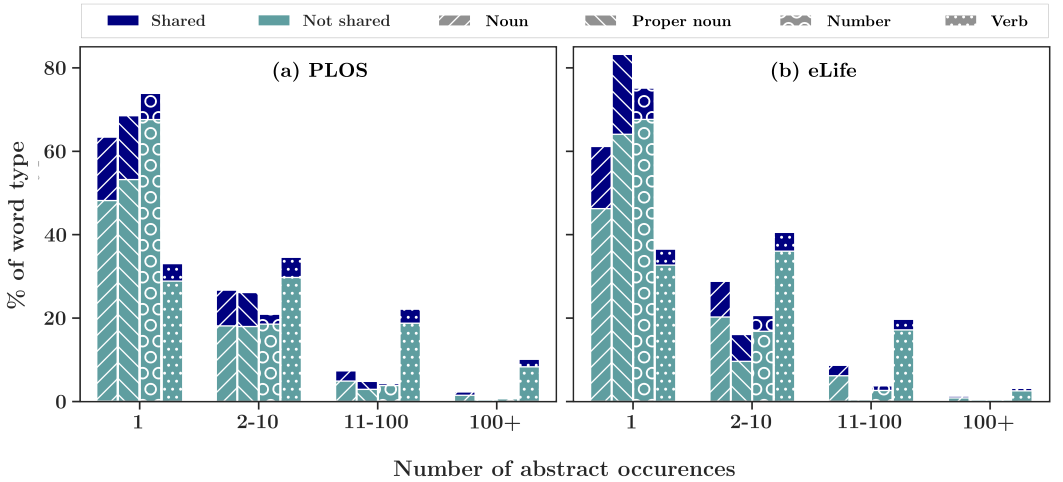}
    }
    \caption{Stacked barplot showing how regularly (on average) abstract content words are shared with the respective lay summaries (as a \% of all words of that type), separated by number of abstract occurrences.}
    \label{fig:content_word_sharing}
\end{figure*}

\subsection{Rhetorical Structure}  \label{subsec:analysis-rhetoric}
Rhetoric is another important factor \textcolor{black}{when assessing the comprehensibility of a text}.
Specifically, a lay person will require a much larger focus on the background of a scientific article than an expert in order to understand the significance of its findings \citep{elifeDigest}, thus we would expect lay summaries to focus more on such aspects. 

To provide further insight into the structural differences between abstracts and lay summaries, we classify all sentences within each based on their rhetorical status. To do this, we make use of PubMed RTC \citep{Dernoncourt2017-te}, a dataset containing the 20,000 biomedical abstracts retrieved from PubMed, with each sentence labelled according to its rhetorical role (roles: Background, Objective, Methods, Results, Conclusions).
We use PubMed RTC to train the BERT-based sequential classifier introduced by \citet{Cohan2019-ru} due to its strong reported performance (92.9 micro F1-score), before applying this model to lay summary and abstract sentences from our datasets.

Figure \ref{fig:discourse_comp} provides a visualisation of how the frequency of each rhetorical class changes according to the sentence position within our summaries. For each sub-graph, observing the pattern of most frequent labels (tallest bars) across all positions allows us to get an idea of the dominant rhetorical structure. In Table~\ref{tab:discourse_percs}, we further quantify the difference in structure by giving the average percentage of each label present in the different summaries. 

For both datasets, we see a similar pattern when comparing abstract and lay summary distributions.
Specifically, a much greater portion of lay summary sentences is dedicated to explaining the relevant background information (``Background"). This is unsurprising, as such information is essential to understanding the motivation and significance \textcolor{black}{of} any work and, thus, would be of great value \textcolor{black}{to a} non-expert. This additional focus on ``Background" comes at the expense of sentences focusing on ``Results" \textcolor{black}{and (to a lesser extent) "Methods"}, which are less frequent within lay summaries. 
\textcolor{black}{Again, this is to be expected, as these details are less meaningful to an audience without domain expertise.}

\subsection{Content Words}  \label{subsec:analysis-content}
Aiming to determine 
what terminology is
shared between summary types, we analyse the frequency at which \textit{content words} occur simultaneously within abstracts and lay summaries.  
We treat nouns, proper nouns, verbs, and numbers as content words, and we extract these from the summaries using ScispaCy \citep{neumann-etal-2019-scispacy}, a library that specialises in the processing of biomedical texts.\footnote{\texttt{en\_core\_sci\_scibert} model used for POS tagging, derived from SciBERT \citep{Beltagy2019-yc}.} 
Figure \ref{fig:content_word_sharing} shows the results of our analysis, visualising the average rate at which different types of content word from the abstracts are shared with lay summaries.
We divide our analysis based on the number of abstracts content words occur in, allowing us to observe how the ubiquity of a word within the corpus affects the rate at which it is shared.
\footnote{E.g., For nouns within \textsc{PLOS}, 15.3\% occur in a single abstract and are shared with lay summaries, and 48.2\% occur in a single abstract and are not shared (i.e., total percentage of nouns that occur in a single abstract = 15.3 + 48.2 = 63.5\%).}

Generally, we observe similar patterns for content words between datasets. 
Firstly, regardless of word type or number of abstract occurrences, we find that abstract content words are rarely shared with lay summaries (i.e., `shared' \% < `not shared' \% for all bars).
This is indicative of a clear shift in content and/or vocabulary when it comes to the creation of a lay summary.  
For each dataset, we can also see that the vast majority of content words of all types, except verbs, occur in 10 or fewer abstracts (> 90\% on average for both datasets), with most of these occurring within a single abstract. 
We found that the majority of these low-frequency content words are highly specific to the topic of a single article or small group of articles. In the case of nouns and proper nouns, we empirically observed these instances to often be highly technical terms (e.g., specific chemicals, lesser-known diseases, etc.), whereas numbers are typically exact numerical figures. It is understandable, therefore, that single-use content words of these types (noun, proper noun, and number) are rarely included in the lay summary, as they will likely be \textcolor{black}{meaningless} to a lay reader. 
The pattern exhibited by verbs differs significantly from that of other word types, as they typically occur in a greater number of abstracts (most commonly being present within 2-10).
For content words of all types, we observe that the ratio of `shared' to `not shared' generally increases in line with the number of abstract occurrences.\footnote{Better illustrated by Table \ref{tab:rare_words_type} in Appendix \ref{sec:appendix1}, which gives the exact percentages shown in Figure \ref{fig:content_word_sharing}.}


\subsection{Abstractiveness}  \label{subsec:analysis-abstractiveness}
\begin{figure}[t]
    \centering
    \resizebox{\columnwidth}{!}{%
        \includegraphics{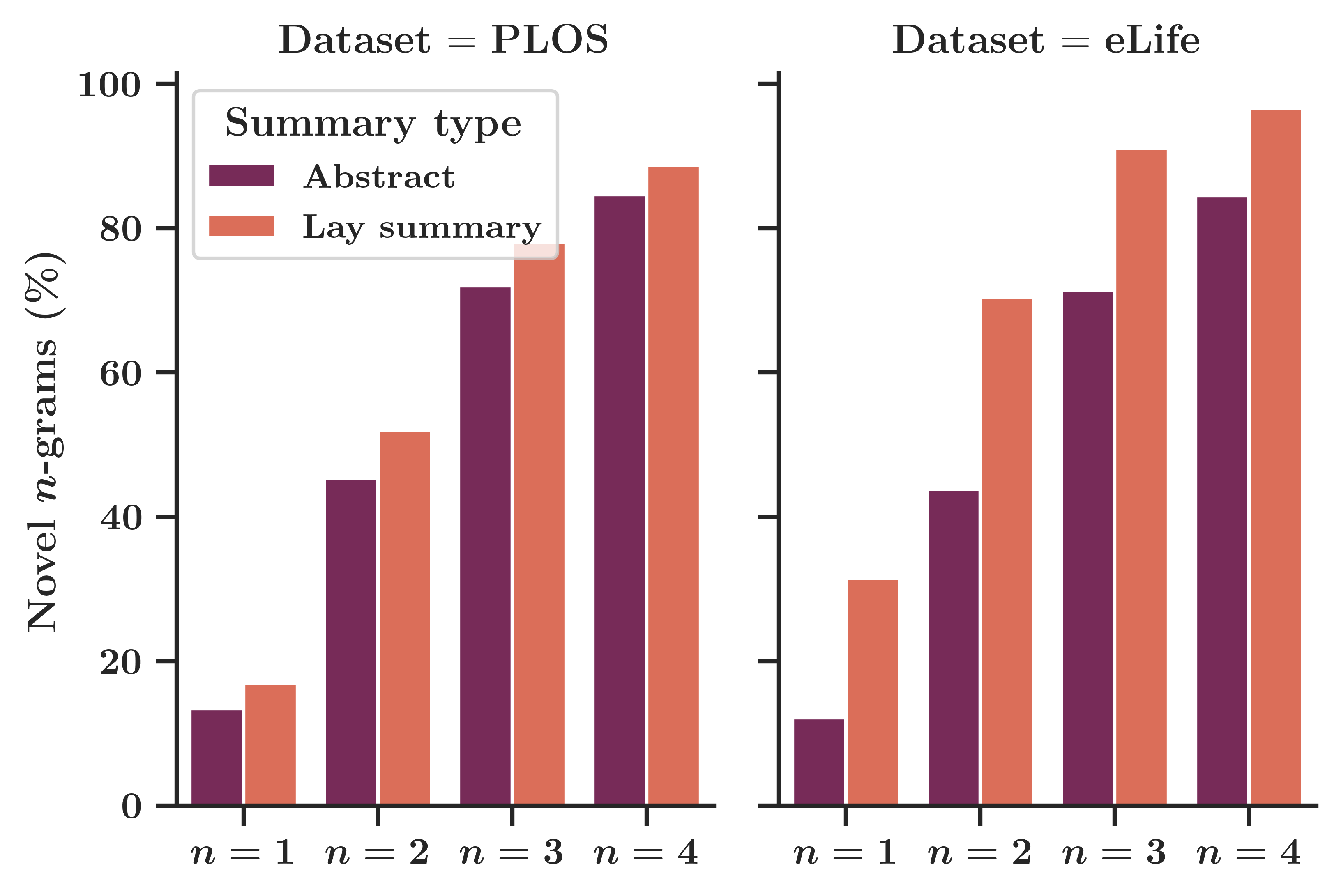}
    }
    \caption{Barplot showing the percentage of novel $n$-grams for each summary type.}
    \label{fig:novel_ngrams}
\end{figure}

We follow the example of prior works \citep{Sharma2019-ue, See2017-ve} 
by calculating the \textit{abstractiveness} of our summaries using $n$-gram novelty,
\textcolor{black}{thus providing a measurement of the degree to which the summary uses different language to describe the content of the article.}
Specifically, for both abstracts and lay summaries, we compute the percentage of summary $n$-grams which are absent from their respective article. The results of this analysis are presented in Figure \ref{fig:novel_ngrams}, where we can \textcolor{black}{observe} that lay summaries consistently contain more novel $n$-grams than abstracts across both datasets. However, the lay summaries of eLife, in addition to being approximately twice as long as those of PLOS (Table \ref{tab:datasets}), appear to be significantly more abstractive. 
Alongside differences in readability (highlighted in \S\ref{subsec:analysis-readability}), we believe these to be important distinctions that should be considered in determining a suitable dataset for a particular use case or application.

\section{Experiments and Results} \label{sec:exp}

To help facilitate future work, we benchmark our datasets using popular heuristics-based, unsupervised, and supervised summarisation approaches  (\S\ref{subsec:results-baselines}). Additionally, we provide further insight into these results via a detailed discussion (\S\ref{subsec:results-discussion}) and an expert-based manual evaluation (\S\ref{subsec:results-human}).   

\begin{table*}[ht]
    \centering
    \resizebox{1.0\textwidth}{!}{
    \begin{tabular}{llccccccccccc}
        \hline \multirow{2}{*}{\textbf{Approach}} && \multicolumn{5}{c}{\textbf{PLOS}} && \multicolumn{5}{c}{\textbf{eLife}}  \\ \cline{3-7} \cline{9-13}
        & & \textbf{R-1} & \textbf{R-2} & \textbf{R-L} & \textbf{FKGL} & \textbf{DCSR}  && \textbf{R-1} & \textbf{R-2} & \textbf{R-L} & \textbf{FKGL} & \textbf{DCSR}  \\
        \hline
           \parbox[t]{1mm}{\scriptsize \multirow{4}{*}{\rotatebox[origin=c]{90}{Heuristic}}} 
           \hspace{3pt} \textsc{Lead-3}      && 25.46 & 6.35  & 22.89 & 15.08 & 12.66 && 17.93 & 3.66 & 16.45 & 13.30 & 12.65  \\
           \hspace{8pt} \textsc{Lead-K}      && 35.50 & 8.69  & 32.33 & 14.94 & 11.88 && 34.12 & 6.73 & 32.06 & 11.89 & 10.58 \\
           \hspace{8pt} \textsc{Abstract}    && 47.07 & 18.60 & 43.51 & 14.98 & 11.10 && 28.92 & 6.19 & 27.04 & 15.35 & 11.87 \\ 
           \hspace{8pt} \textsc{Oracle\_Ext} && 56.43 & 31.24 & 52.88 & 15.28 & 11.20 && 46.38 & 11.48 & 43.82 & 13.18 & 10.51 \\ \hline
           \parbox[t]{1mm}{\scriptsize \multirow{4}{*}{\rotatebox[origin=c]{90}{Unsupervised}}}
            \hspace{3pt} \textsc{LSA} && 35.14 & 6.92 & 31.46 & 17.39 & 12.41 && 34.57 & 5.69 & 32.27 & 16.50 & 11.52 \\            
           \hspace{8pt} \textsc{LexRank}  && 36.53 & 9.43 & 33.09 & 14.90 & \textbf{10.51} && 30.32 & 5.29 & 28.40 & 13.45 & 9.75  \\
           \hspace{8pt} \textsc{TextRank}  && 36.03 & 8.71 & 32.04 & 20.42 & 11.83 && 31.45 & 5.28  & 29.22 & 19.13 & 11.28 \\
           \hspace{8pt} \textsc{HipoRank}  && 42.16 & 11.41 & 38.01 & 14.91 & 11.82 && 29.95 & 5.38 & 27.44 & 13.58 & 12.32 \\ \hline
           \parbox[t]{1mm}{\scriptsize \multirow{4}{*}{\rotatebox[origin=c]{90}{Supervised}}} 
           \hspace{3pt} \textsc{Bart}  && \textbf{42.35} & 12.96 & 38.57 & 14.62 & 12.07  && \textbf{46.57} & \textbf{11.65} & \textbf{43.70} & 10.94 & 9.36  \\
            \hspace{8pt} \textsc{Bart$_{PubMed}$}  && 42.12 & 12.70 & 38.34 & 14.75 & 12.09 && 46.20 & 11.36 & 43.21 & 11.24 & 9.64 \\
           \hspace{8pt} \textsc{Bart$_{Cross}$}  && 42.24 & \textbf{13.52} & \textbf{38.63} & \textbf{14.24} & 12.18  &&  46.22 & 11.53 & 43.33 & 10.81 & 9.31 \\
           \hspace{8pt} \textsc{Bart$_{Scaffold}$}  && 39.47 & 9.73 & 35.79 & 14.45 & 11.74 && 45.28 & 10.99 & 42.51 & \textbf{10.65} & \textbf{9.19} \\ 
           \hline
    \end{tabular}
    }
    \caption{Performance of summarisation models on the test splits of each dataset (\textbf{R} = average ROUGE F1-score). The best non-heuristic scores for each metric are given in \textbf{bold}.}
    \label{tab:summ_results}
\end{table*}

\subsection{Baseline Approaches} \label{subsec:results-baselines}
For our heuristics-based approaches, we include the widely-used \textsc{Lead-3} baseline which simply uses the first three sentences of the main body of the text. As our lay summaries typically consist of more than three sentences, we also include \textsc{Lead-k},
with $k$ being equal to the average lay summary length for each dataset (Table \ref{tab:datasets}). Additionally, we include the scores obtained by the technical abstracts (\textsc{Abstract}) and \textcolor{black}{\textsc{Oracle\_Ext}, a greedy extractive oracle \citep{Nallapati2017}
that provides an upper bound for the expected performance of extractive models.}\footnote{The extractive oracle is further explained in Appendix \ref{sec:appendix1}.}

We benchmark four unsupervised extractive approaches: \textsc{LSA} \citep{Steinberger2004}, \textsc{LexRank} \citep{Erkan_2004}, \textsc{TextRank} \citep{mihalcea-tarau-2004-textrank}, and \textsc{HipoRank} \citep{Dong2021-yh}.
\textcolor{black}{For supervised models, we use the transformer-based \textsc{BART} base model \citep{lewis-etal-2020-bart}, which we fine-tune on our own datasets.} 
To assess how the use of additional data in various forms can benefit performance, we include several other variants of this model which are described in the remainder of this subsection.\footnote{For all variants, we truncate the input to 1,024 tokens.}

\paragraph{Additional training} 
As our datasets remain smaller than those used in other forms of summarisation, \textcolor{black}{we experiment with} \textsc{BART}$_{PubMed}$, which is previously trained on the PubMed abstract generation dataset \citep{Cohan2018-nq} and fine-tuned on our own datasets. Aiming to assess how well models trained on PLOS can generalise to eLife and vice versa, we also include \textsc{BART}$_{Cross}$, which is initially trained on the opposite dataset to that which it is eventually fine-tuned and evaluated on.

\paragraph{Scaffolding}
We also experiment with artificially enlarging our training data by way of a scaffold task. Inspired by CATTS \citep{Cachola2020-bx}, we remove the article's abstract from the input text and train the model to generate the abstract as a scaffold task to lay summarisation. In addition to showing whether training for abstract generation can benefit lay summarisation, results for this model will provide an indication of the baseline \textsc{BART} model's reliance on the abstract content.
Specifically, we include two copies of every article within our training data - one using the abstract as the reference summary and the other using the lay summary. We distinguish between the two by prepending the input document with the control tokens \texttt{$\langle|$ABSTRACT$|\rangle$} or \texttt{$\langle|$SUMMARY$|\rangle$}. Documents within the validation and test splits are prepended the \texttt{$\langle|$SUMMARY$|\rangle$} code to induce lay summary generation. This model is denoted by \textsc{BART}$_{Scaffold}$. 

\subsection{Discussion} \label{subsec:results-discussion}
Table \ref{tab:summ_results} presents the performance of the aforementioned approaches on the PLOS and eLife test splits using automatic metrics. In line with common practice for summarisation, we report the F1-scores of ROUGE-1, 2, and L \citep{lin-2004-rouge}. Additionally, we include FKGL and DCRS scores of the generated output (see \S\ref{subsec:analysis-readability}), providing an assessment of the syntactic and lexical complexity, respectively. 

\paragraph{The importance of the abstract}
Based on the ROUGE scores obtained by the \textsc{Abstract} baseline, we can safely assume that the lay summaries of PLOS are much closer in resemblance to their respective abstracts than those of eLife. 
The importance of the abstract for lay summary generation is further highlighted by the ROUGE scores of the \textsc{BART}$_{Scaffold}$ model, which performs notably worse than the standard \textsc{BART} model on PLOS and slightly worse on eLife. These results suggest that having the abstract included within the model input provides significantly more benefit than using abstract generation as an auxiliary training signal.

\paragraph{Extractive vs abstractive}
In general, we would expect abstractive methods to have greater application for the task of lay summarisation due to their ability to transform (and thus, simplify) an input text. However, abstractive approaches have a tendency to generate hallucinations, resulting in factual inconsistencies between the source and output that damages their usability \citep{Maynez2020-xd}. Therefore, extractive approaches may still have utility for the task, especially if the comprehensibility of selected sentences is directly considered. 

For ROUGE scores, we find that extractive baselines (i.e., all unsupervised and heuristic approaches) perform significantly better on PLOS than on eLife, aligning with our previous analysis (\S\ref{subsec:analysis-abstractiveness}) which identified PLOS as the less abstractive dataset.
Interestingly, readability scores achieved by extractive models on PLOS match and sometimes exceed those of abstractive \textsc{BART} models, although they are inconsistent. 
For eLife, abstractive methods (i.e., \textsc{BART} models) generally obtain superior scores for both ROUGE and readability metrics. In fact, the ROUGE scores achieved by \textsc{BART} exceed those obtained by \textsc{Oracle\_Ext}, further indicating that abstractive methods have greater potential for this dataset.

%

\paragraph{Use of additional data}
As previously mentioned, artificially creating more data via an abstract-generation scaffold task results in a decrease in ROUGE scores for both datasets, indicating a reliance on the abstract content for lay summarisation.  We also find that pretraining \textsc{BART} on Pubmed \textcolor{black}{(\textsc{BART}$_{PubMed}$)} does very little to affect the performance, suggesting that habits learned for abstract generation do not transfer well to lay summarisation.
Similarly, \textsc{BART}$_{Cross}$ achieves a performance close to that of the 
\textcolor{black}{standard \textsc{BART} model.}
Overall, these results indicate that additional out-of-domain training does provide much benefit for lay summarisation, and alternative modelling approaches that make better use of available data may be a more promising route for future work. 

\subsection{Human evaluation} \label{subsec:results-human}
To further assess the usability of the generated abstractive summaries, we perform an additional human evaluation of our standard \textsc{Bart} baseline model using two domain experts.\footnote{Both judges have experience in scientific research and hold at least a bachelor's degree in Biomedical Science.} 

Our evaluation uses a random sample of 10 articles from the test split of each dataset. Alongside each model-generated summary, judges are presented with both the abstract and reference lay summary of the given article. Using a 1-5 \textit{Likert} scale, the annotators are asked to rate the model output based on three criteria:
(1) \textit{Comprehensiveness} -- to what extent does the model output contain all information that might be necessary for a non-expert to understand the high-level topic of the article and the significance of the research;
(2) \textit{Layness} -- to what extent is the content of the model output comprehensible (or readable) to a non-expert, in terms of both structure and language;
(3) \textit{Factuality} -- to what extent is the model output factually consistent with the two other provided summaries. 
We choose not to provide judges with the full article text in an effort to minimise the complexity of the evaluation and the cognitive burden placed upon them.

\begin{table}[]
    \centering
    \begin{tabular}{cccc}
         \hline
         \textbf{Dataset} & \textbf{Comp.} & \textbf{Layness} &\textbf{Factuality} \\ \hline
         PLOS & 3.7 & 3.0 & 3.0 \\
         eLife & 3.1 & 3.0 & 3.0 \\ \hline 
    \end{tabular}
    \caption{Mean evaluator ratings (1-5) obtained by \textsc{Bart} outputs for each metric.}
    \label{tab:human_eval}
\end{table}

Table \ref{tab:human_eval} presents the average ratings from our manual evaluation. We calculate Krippendorff's $\alpha$ to measure inter-rater reliability, where we obtain values of 0.78 and 0.54 for PLOS and eLife, respectively. In addition to providing ratings, evaluators also provided comments on the general performance on each criterion for both datasets, providing further insights into model performance.

\paragraph{Comprehensiveness}

We can see from Table \ref{tab:human_eval} that model outputs on PLOS are judged to be more comprehensive than on eLife. From evaluators' comments, we understand that this largely results from extensive use of abstract content for PLOS, which is sometimes copied directly (or with minor edits) to the lay summary.  
For eLife, it was observed that new information (i.e., not contained in the reference abstract or lay summary) was often introduced which was irrelevant or confusing, potentially affecting the understanding of a lay reader.

\paragraph{Layness}
Interestingly, given the previously highlighted differences in readability, the average layness of the model output is judged to be equal for both datasets (3.0), \textcolor{black}{suggesting a reasonable degree of content simplification.}  
However, evaluators' comments indicate that model outputs for each dataset were penalised for different reasons. For PLOS, the aforementioned use of abstract content often resulted in the inclusion of jargon terms that a lay reader would struggle to interpret. Alternatively, the language of eLife outputs was observed to be better suited to a lay audience but was sometimes simplified to a point that it could be \textcolor{black}{misconstrued and mislead a reader, occasionally containing grammatical errors, typos, and repeated content.} 


\paragraph{Factuality}

Again, we find an equal average score of 3.0 given for factuality, suggesting the model struggles to produce factually correct outputs for both datasets. In fact, we found no output from either dataset was given a perfect score by both annotators, indicating that simplifying technical content accurately is a consistent problem. Evaluators' comments highlight contradictions, unclear phrasing, and misrepresentation of entities as key contributing factors to factual inconsistencies. We regard this as an integral obstacle to overcome in the development of usable lay summarisation models and an essential focus for future research.





\section{Conclusion} \label{sec:conclusion}
In this work, we have introduced PLOS and eLife, two new datasets for the lay summarisation of biomedical research articles. 
Compared to currently available resources, these datasets possess source article and summary formats that are more broadly applicable to wider literature, with PLOS also being larger by a significant margin.    
A thorough analysis of our lay summaries highlights key differences  between datasets, enabling them to cater to the needs of different audiences and applications. Specifically, in addition to being approximately twice as long as those of PLOS, we find eLife summaries to be both more readable and abstractive, thus better suited to a less technical audience.
To facilitate future research, we benchmark our datasets with popular summarisation models using automatic metrics and conduct an expert-based human evaluation, providing further insight into the intricacies of model performance on our datasets and highlighting key challenges for the task of lay summarisation.

\section{Limitations}
\label{sec:limitations}
Although we introduce the largest dataset available to the task of lay summarisation, our datasets remain smaller than those available for other forms of summarisation (e.g., abstract generation), where there exists datasets containing 100,000+ articles. This is largely due to the fact that lay summaries are less ubiquitous that other forms of summary (e.g., the abstract), only being used in a relatively small number of journals, of which only some are open-access and available to be utilised for such purposes as dataset creation. 

On a related note, another potential limitation of our datasets is the fact they only cover a single broad domain - biomedicine. Again, this comes down to the availability of data, and the fact that the use of lay summaries is much less common in other scientific domains (e.g. Computer Science).
There is, however, a reason for this disparity in the adoption of lay summaries between domains, as it is generally considered more important that the public have an awareness and understanding of research breakthroughs in health-related areas such as biomedicine.
Therefore, we believe it is in these domains that automatic lay summarisation can provide the greatest benefit, although we also hope to address the lay summarisation of other domains in future work.

\section{Acknowledgements}

This work was supported by the Centre for Doctoral Training in Speech and Language Technologies (SLT) and their Applications funded by UK Research and Innovation [grant number EP/S023062/1].

\bibliography{anthology,custom}
\bibliographystyle{acl_natbib}

\appendix
\section{Appendix} \label{sec:appendix1}

\paragraph{Availability of data sources}
Both PLOS and eLife are open-access journals, with an emphasis on making scientific research accessible to a wide audience. PLOS articles are available to be mined, reused, and shared by anyone, as per their data mining policy.\footnote{\url{https://plos.org/text-and-data-mining}} eLife articles are available under the permissive CC-BY 4.0 license, and thus also available to be retrieved and shared for these purposes.\footnote{\url{https://elifesciences.org/inside-elife/6933fe8e/resources-for-developers}} The data for PLOS and eLife was retrieved on 7/03/22 and 11/03/22, respectively. All articles and summaries are in English only. Our datasets are made available to the community to facilitate future research.

\paragraph{Additional data processing details}
Here we provide some additional details regarding the dataset creation process, building on the description given in \S\ref{sec:our-datasets}. 
For eLife, the XML files retrieved were found to include multiple versions of the same article, identifiable by the article id which includes a version number. For these, we removed duplicates and kept only the most recent versions. 

For both datasets, prior to extraction, we remove all Tables, Figures, and sections marked with the tag ``supplementary-material". We also do not extract sections with the heading ``acknowledgments". 
During sentence segmentation, we use a regular expression to identify and temporarily replace all "et al." occurrences with unique placeholder tokens, which are then replaced following segmentation. 
Following segmentation, we again use a regular expression to identify and remove all sentences which began with ``DOI:" followed by a URL from both abstracts and lay summaries, as these were found to commonly occur at the end of both.

\begin{figure*}[ht]
    \centering
    \resizebox{0.8\textwidth}{!}{%
        \includegraphics{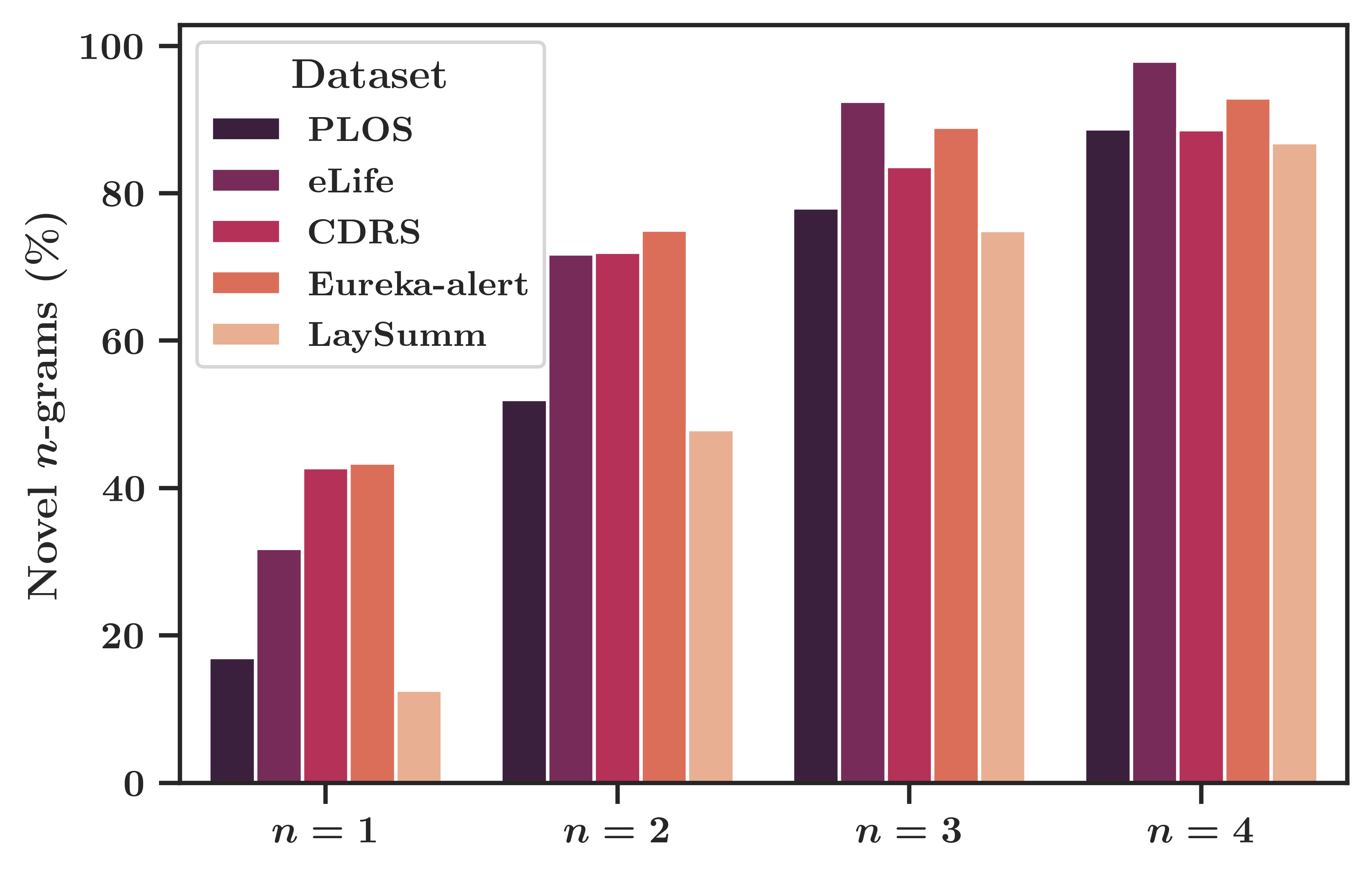}
    }
    \caption{Comparison of lay summary $n$-gram novelty for all lay summarisation datasets.}
    \label{fig:ds_novelty_comp}
\end{figure*}

\paragraph{Comparison to previous datasets} 
We provide a comparison of the readability and abstractiveness of lay summaries for all lay summarisation datasets in Table \ref{tab:dataset_readability} and Figure \ref{fig:ds_novelty_comp}, respectively.

\paragraph{Supplementary figure details}
Table \ref{tab:rare_words_type} gives the exact percentages that are visualised in Figure \ref{fig:content_word_sharing}, allowing for a more detailed analysis of content word sharing (e.g., calculating the ratio `shared' to `not shared' for different content word types).

\paragraph{Baseline model details}
Here we provide additional experimental details for our baselines approaches (Table \ref{tab:summ_results}). 
\textsc{Oracle\_Ext} is a greedy oracle, which means it repeatedly extracts the next article sentence that will maximise the mean ROUGE scores (1, 2, and L) of the extracted summary, up to the maximum length (equal to the average lay summary length for a given dataset - Table \ref{tab:datasets}).

For all \textsc{BART} models, we make use of the \texttt{huggingface} library \citep{huggingface}. Specifically, we use the ``facebook/bart-base" model for baselines \textsc{BART}, \textsc{BART$_{Cross}$}, and \textsc{BART$_{Scaffold}$}, and we use the ``mse30/bart-base-finetuned-pubmed" model for \textsc{BART$_{PubMed}$}. Training was run (using 4x NVIDIA Tesla V100 SXM2 GPUs) for all models with AdamW optimisation \citep{Loshchilov2019DecoupledWD} and an early stopping patience of 25 epochs, with the best model being selected by performance on the validation set (ROUGE-2).

All unsupervised baselines were run with default configurations.

\paragraph{Automatic evaluation}
For the calculation of ROUGE scores, we use the \texttt{rouge-score} Python package.\footnote{\url{https://pypi.org/project/rouge-score/}} For FKGL and DCSR, we use the \texttt{textstat} Python package.\footnote{\url{https://github.com/shivam5992/textstat}} 

\paragraph{Human evaluation comments}
Comments on the general model performance for each criterion provided by each annotator for our human evaluation are given in Figures \ref{fig:plos_eval_comments} and  \ref{fig:elife_eval_comments} for PLOS and eLife, respectively. 

\paragraph{Lay summary examples}
Full examples of lay summaries and their respective technical abstracts are given in Figures \ref{fig:laysum_example_plos} and \ref{fig:laysum_example_elife} for PLOS and eLife, respectively.

\begin{table*}[]
    \centering
    \begin{tabular}{lcccc}
        \hline
        \textbf{Dataset} & \textbf{FKGL} & \textbf{CLI} & \textbf{DCRS} & \textbf{WordRank}  \\
        \hline
        \textsc{LaySumm}        & $14.81_{\pm 2.91}$ & $16.28_{\pm 2.81}$ & $11.63_{\pm 1.16}$ & $9.04_{\pm 0.60}$ \\
        \textsc{Eureka-Alert}   & $13.16_{\pm 2.07}$ & $14.09_{\pm 1.56}$ & $9.77 _{\pm 0.85}$ &  $8.79_{\pm 0.80}$ \\
        \textsc{CDSR}           & $12.72_{\pm 2.16}$ & $14.17_{\pm 0.88}$ & $9.073_{\pm 0.88}$ & $8.54_{\pm 0.32}$ \\
        \textbf{\textsc{PLOS}}  & $14.76_{\pm 2.33}$ & $15.90_{\pm 2.01}$ & $10.91_{\pm 0.85}$ & $8.98_{\pm 0.32}$ \\
        \textbf{\textsc{eLife}} & $10.91_{\pm 1.44}$ & $12.52_{\pm 1.35}$ & $8.94_{\pm 0.53}$ & $8.68_{\pm 0.31}$ \\
        \hline
    \end{tabular}
    \caption{Comparison of the lay summary readability scores for all lay summarisation datasets.}
    \label{tab:dataset_readability}
\end{table*}

\begin{table*}[t]
    \centering
    \begin{tabular}{llcccc} \hline
        \multirow{2}{*}{\textbf{Word type}} && \multicolumn{4}{c}{\textbf{\# of abstract occurrences}}  \\ 
        \cline{3-6} && \textbf{1} & \textbf{2-10} & \textbf{11-100} & \textbf{100+} \\ 
        \hline
        \parbox[t]{1mm}{\footnotesize \multirow{4}{*}{\rotatebox[origin=c]{90}{\textbf{PLOS}}}}
        \hspace{3pt} Noun        && 15.3 / 48.2 & 8.7 / 18.1 & 2.5 / 4.9  & 0.8 / 1.5 \\
        \hspace{8pt} Proper noun && 15.3 / 54.2 & 8.2 / 18.0 & 1.9 / 2.9  & 0.2 / 0.3 \\   
        \hspace{8pt} Number      && 6.3  / 67.6 & 2.4 / 18.6 & 0.5 / 3.8  & 0.2 / 0.6 \\ 
        \hspace{8pt} Verb        && 4.2 / 28.9  & 4.9 / 29.7 & 3.4 / 18.9 & 1.8 / 8.3 \\
        \hline
        \parbox[t]{1mm}{\footnotesize \multirow{4}{*}{\rotatebox[origin=c]{90}{\textbf{eLife}}}}
        \hspace{3pt} Noun        && 14.9 / 46.2 & 8.6 / 20.3 & 2.5 / 6.2 & 0.5 / 0.9 \\
        \hspace{8pt} Proper noun && 19.2 / 64.1 & 6.4 / 9.6  & 0.3 / 0.4 & 0.0 / 0.0 \\
        \hspace{8pt} Number      && 7.6  / 67.6 & 3.8 / 16.8 & 1.2 / 2.6 & 0.2 / 0.2 \\ 
        \hspace{8pt} Verb        && 3.8  / 32.7 & 4.5 / 36.0 & 2.6 / 17.2 & 0.6 / 2.6  \\
        \hline
\end{tabular}
    \caption{Statistics for bars in Figure \ref{fig:content_word_sharing}. For each table cell, the overall percentage is split into `\% that are shared with lay summaries' / `\% that are not shared with lay summaries'.}
    \label{tab:rare_words_type}
\end{table*}

\begin{figure*}
  \begin{minipage}{\textwidth}
    \noindent\fbox{%
        \parbox{\textwidth}{%
            \textbf{Comprehensiveness} \\ 
            \textit{Annotator 1:} The model outputs summarised the important information, however it also cut out quite a lot of background info which is key for understanding the science. Overall, the comprehensiveness was enough for a non-expert to grasp the overall gist of the studies.
            
            \textit{Annotator 2:} The model seems to use parts of the abstract, and therefore seems quite comprehensive. It also does a decent job of a final ``summary" sentence to the lay summary to summarize/put into context. 
        }%
    }
    \noindent\fbox{%
        \parbox{\textwidth}{%
            \textbf{Layness} \\
            \textit{Annotator 1:} Some abstracts contained a lot of jargon which would be confusing/off putting to a non-expert. Although I know some scientific words cannot be substituted, it would be good to have an explanation of the more complex words in brackets, for example. 
            
            \textit{Annotator 2:} Due to the overlap with the reference abstract, the output is comprehensive but probably confusing to a lay audience, in some cases there is no introduction/background on scientific jargon (e.g. we cannot expect a lay audience to understand complex scientific techniques, cellular or molecular machinery). Use of Genus species nomenclature is also likely to confuse lay audiences, where a common name could be used instead, not as well as (e.g. `Egyptian mosquito' instead of `Aedes aegypti', also known as the Egyptian mosquito'). 
        }%
    }
    \noindent\fbox{%
        \parbox{\textwidth}{%
            \textbf{Factuality} \\
            \textit{Annotator 1:} The majority of the statements were factually correct, although sometimes the meaning of the simplified language could be misinterpreted, which would result in a similar outcome to factually wrong statements.
            
            \textit{Annotator 2:} Some minor factual errors throughout, and mixups between gene symbols (e.g. where one letter will be changed, PMN to PMA), there are also come cases where it will pull out a \% but mix up what gene / condition it is related to, ultimately leading to the formation of a sentence which is factually incorrect.
        }%
    }
  \end{minipage} \quad
    \caption{Human evaluation comments for PLOS.}
  \label{fig:plos_eval_comments}
\end{figure*}

\begin{figure*}
  \begin{minipage}{\textwidth}
    \noindent\fbox{%
        \parbox{\textwidth}{%
            \textbf{Comprehensiveness} \\
            \textit{Annotator 1:} Overall the information contained within the model-generated summaries effectively conveyed the information in the references. However, there was a few occasions where new elements/concepts were introduced that could confuse the reader and affect their understanding (these were sometimems factual statements, sometimes seemingly made up). The model abstract did provide enough information for a general understanding of the topic and would be sufficient as a brief overview. 
            
            \textit{Annotator 2:} The model usually picks up on the core points of the abstract but can often introduce extra information which is either off-topic or factually incorrect. The model seems to start off by introducing the topic well but struggles to hit the ``what is the significance of this research?" question.
        }%
    }
    \noindent\fbox{%
        \parbox{\textwidth}{%
            \textbf{Layness} \\
            \textit{Annotator 1:} The language was, in the majority, well suited to a lay person and terminology was adapted accordingly. However, at times the information was simplified to a point where it could be misconstrued which, with scientific information, is a potential risk. At times, jargon still remained and I could imagine some people being confused by this. There were a few grammatical errors and poor sentence structure, typos, repetition etc. 
            
            \textit{Annotator 2:} Sometimes the model introduces extra information which is not suitable for a lay audience, for example: references to genome sequencing, progenitor cells, endoplasmid reticulum. There are also instances where misinterpretation by the model may mislead the lay audience, for example there was an output where Norepinephrine was said to be ``a.k.a. dopamine", which is not factually correct. 
        }%
    }
    \noindent\fbox{%
        \parbox{\textwidth}{%
            \textbf{Factuality} \\
            \textit{Annotator 1:} There were a few summaries which contained incorrect information, things that are well-known in the scientific community were poorly conveyed. At times, new information was introduced which contradicted earlier statements, and those of the reference abstracts/lay summaries. Of course, some information was correct. I would be concerned about the level of misinformation which could arise from these summaries, if used to educate a lay audience.
            
            \textit{Annotator 2:} This seemed to vary based on the abstract and how well the output started, for example if the model introduced the topic well, it would lead to more factual points. However, there were some generated summaries which were factually incorrect from the start and this lead to more errors.
        }%
    }
  \end{minipage} \quad
    \caption{Human evaluation comments for eLife.}
  \label{fig:elife_eval_comments}
\end{figure*}

\begin{figure*}
  \begin{minipage}{\textwidth}
    \noindent\fbox{%
        \parbox{\textwidth}{%
            \textbf{Technical Abstract} \\
            Rabies is a uniformly fatal disease, but preventable by timely and correct use of post exposure prophylaxis (PEP). Unfortunately, many health care facilities in Pakistan do not carry modern life-saving vaccines and rabies immunoglobulin (RIG), assuming them to be prohibitively expensive and unsafe. Consequently, Emergency Department (ED) health care professionals remain untrained in its application and refer patients out to other hospitals. The conventional Essen regimen requires five vials of cell culture vaccine (CCV) per patient, whereas Thai Red Cross intradermal (TRC-id) regimen requires only one vial per patient, and gives equal seroconversion as compared with Essen regimen. This study documents the cost savings in using the Thai Red Cross intradermal regimen with cell culture vaccine instead of the customary 5-dose Essen intramuscular regimen for eligible bite victims. All patients presenting to the Indus Hospital ED between July 2013 to June 2014 with animal bites received WHO recommended PEP. WHO Category 2 bites received intradermal vaccine alone, while Category 3 victims received vaccine plus wound infiltration with Equine RIG. Patients were counseled, and subsequent doses of the vaccine administered on days 3, 7 and 28. Throughput of cases, consumption utilization of vaccine and ERIG and the cost per patient were recorded. Government hospitals in Pakistan are generally underfinanced and cannot afford treatment of the enormous burden of dog bite victims. Hence, patients are either not treated at all, or asked to purchase their own vaccine, which most cannot afford, resulting in neglect and high incidence of rabies deaths. TRC-id regimen reduced the cost of vaccine to 1/5th of Essen regimen and is strongly recommended for institutions with large throughput. Training ED staff would save lives through a safe, effective and affordable technique.
        }%
    }
    \noindent\fbox{%
        \parbox{\textwidth}{%
            \textbf{Lay Summary} \\
            Rabies is a killer disease caused by the rabies virus that is present in the saliva of rabid animals, mainly the dog. Once symptoms become apparent, death is inevitable. However, rabies can be prevented if correct post exposure prophylaxis (PEP) is instituted as soon as possible after a dog bite and before symptoms of rabies begin. In Pakistan, government hospitals treat 50-70 new bite victims each day. Many still dispense the free but poor quality nerve tissue vaccine that is often ineffective and fraught with serious adverse reactions. Hospital administrators consider PEP too expensive to be administered free of cost. The Indus Hospital (TIH), Karachi is a private teaching hospital which provides free medical care to all. From July 2013-June 2014, 2,983 new bites were seen in the ED, and rather than use the Essen regimen of five full vial intramuscular doses per patient over 28 days, we administered the WHO-approved Thai Red Cross-intradermal (TRC-id) 4-dose regimen. The use of the TRC-id regimen resulted in 80\% cost savings over the Essen regimen. In resource-poor settings, we advocate training of ED personnel in TRC-id regimen, which, ultimately, will result in less vaccine consumption, more patient compliance and complete treatment, resulting in more lives being saved.
        }%
    }
  \end{minipage} \quad
    \caption{PLOS lay summary example.}
  \label{fig:laysum_example_plos}
\end{figure*}

\begin{figure*}
  \begin{minipage}{\textwidth}
    \noindent\fbox{%
        \parbox{\textwidth}{%
            \textbf{Technical Abstract} \\
            Adult stem cells are responsible for life-long tissue maintenance. They reside in and interact with specialized tissue microenvironments (niches). Using murine hair follicle as a model, we show that when junctional perturbations in the niche disrupt barrier function, adjacent stem cells dramatically change their transcriptome independent of bacterial invasion and become capable of directly signaling to and recruiting immune cells. Additionally, these stem cells elevate cell cycle transcripts which reduce their quiescence threshold, enabling them to selectively proliferate within this microenvironment of immune distress cues. However, rather than mobilizing to fuel new tissue regeneration, these ectopically proliferative stem cells remain within their niche to contain the breach. Together, our findings expose a potential communication relay system that operates from the niche to the stem cells to the immune system and back. The repurposing of proliferation by these stem cells patch the breached barrier, stoke the immune response and restore niche integrity.
        }%
    }
    \noindent\fbox{%
        \parbox{\textwidth}{%
            \textbf{Lay Summary} \\
            Most, if not all, tissues of an adult animal contain stem cells. These stem cells regenerate and repair damaged tissues and organs for the entire lifetime of an animal, contributing to a healthy life. They divide to make daughter cells that become either new stem cells or specialized cells of that organ. Adult stem cells exist in specific areas within tissues known as niches, where they interact with surrounding cells and molecules that inform their behavior. For example, cells and molecules within these niches can signal stem cells to remain in a `dormant' state, but upon injury, they can mobilize stem cells to form new tissue and repair the wound. So far, it has been unclear how stem cells sense damage and stress and direct their efforts away from their normal duties towards repair. Here, Lay et al. studied the stem cells in the mouse skin that are responsible to regenerate hair. Every hair follicle contains a niche (the `bulge'), where these stem cells live and share their environment with cells that anchor the hair. The niche tethers to the stem cells through specific adhesion molecules that also help the niche to form a tight seal to prevent bacteria from entering. Lay et al. removed one of the adhesion molecules called E-cadherin, which caused a breach in the niche's barrier. The stem cells sensed their damaged niche, prepared to multiply, and sent out stress signals to the immune system. The immune cells then arrived at the niche and sent signals back to the stem cells, prodding them to multiply and patch the barrier, while at the same time, keeping the inflammation in check. This remarkable ability of the stem cells to recruit immune cells and initiate a dialogue with them enabled the stem cells to divert their attention from regenerating hair and instead directing it towards the site of the tissue damage. Other stem cells, such as those in the lung or gut, may have similar mechanisms to detect and respond to physical damage. It will be interesting to investigate the underlying mechanism of how immune cells are involved in balancing stem cell regenerative capacity and response to physical damage. A better knowledge of these processes could help to regenerate tissues or even entire organs.
        }%
    }
  \end{minipage} \quad
    \caption{eLife lay summary example.}
  \label{fig:laysum_example_elife}
\end{figure*}


\end{document}